\documentclass[conference]{IEEEtran}
\IEEEoverridecommandlockouts
\usepackage{cite}
\usepackage{amsmath,amssymb,amsfonts}
\usepackage{algorithmic}
\usepackage{graphicx}
\usepackage{textcomp}
\usepackage{xcolor}
\def\BibTeX{{\rm B\kern-.05em{\sc i\kern-.025em b}\kern-.08em
    T\kern-.1667em\lower.7ex\hbox{E}\kern-.125emX}}
\begin{document}

\title{Sign Language Recognition Based On Facial Expression and Hand Skeleton\\{\footnotesize}
\thanks{\rule[3pt]{8.5cm}{0.05em}
This work was supported by Southeast University Innovation and Entrepreneurship Program (202210286057Z). Zhiyu Long, Xingyou Liu and Jiaqi Qiao are with the School of Automation and Zhi Li is with the School of Electronic Science and Engineering, Southeast University, Nanjing, China, 210096. Zhiyu Long and Xingyou Liu are co-first authors. The authorship was determined by a coin toss.}
}
\author{\IEEEauthorblockN{1\textsuperscript{st} Zhiyu Long}
\IEEEauthorblockA{\textit{the School of Automation} \\
\textit{Southeast University}\\
Nanjing, China \\
zhiyulong@seu.edu.cn}
\and
\IEEEauthorblockN{2\textsuperscript{nd} Xingyou Liu}
\IEEEauthorblockA{\textit{the School of Automation} \\
\textit{Southeast University}\\
Nanjing, China \\
213202691@seu.edu.cn}
\and
\IEEEauthorblockN{3\textsuperscript{rd} Jiaqi Qiao}
\IEEEauthorblockA{\textit{the School of Automation} \\
\textit{Southeast University}\\
Nanjing, China \\
qiaojiaqi@seu.edu.cn}
\and
\IEEEauthorblockN{4\textsuperscript{th} Zhi Li}
\IEEEauthorblockA{\textit{the School of Electronic Science and Engineering} \\
\textit{Southeast University}\\
Nanjing, China \\
213202568@seu.edu.cn}
}

\maketitle

\begin{abstract}
Sign language is a visual language used by the deaf and dumb community to communicate. However, for most recognition methods based on monocular cameras, the recognition accuracy is low and the robustness is poor. Even if the effect is good on some data, it may perform poorly in other data with different interference due to the inability to extract effective features. To solve these problems, we propose a sign language recognition network that integrates skeleton features of hands and facial expression. Among this, we propose a hand skeleton feature extraction based on coordinate transformation to describe the shape of the hand more accurately. Moreover, by incorporating facial expression information, the accuracy and robustness of sign language recognition are finally improved, which was verified on A Dataset for Argentinian Sign Language and SEU's Chinese Sign Language Recognition Database (SEUCSLRD).
\end{abstract}

\section{Introduction}
Sign language usually includes a variety of features, such as hands, arm posture and facial expression, etc. However, most sign language actions are difficult for normal people who have not received sign language training to understand, making it difficult for deaf and dumb people to communicate with normal people. The goal of sign language recognition is to build a bridge for the communication between the deaf people, the dumb people and the normal people, and improve their life. Sign language recognition involves computer vision, artificial intelligence, natural language processing and other fields. Researching sign language recognition technology is not only conducive to promoting the development of multi-domain technology, but also has far-reaching social significance.

In recent years, many studies on sign language recognition have been carried out on data gloves and the Microsoft Kinect depth camera. The equipment is relatively complex and it’s not an appropriate choice for practical real-time applications. The accuracy of vision based model is generally lower than that of sensor based model, so it is very important to adopt robust and effective methods to extract manual and non-manual features. For instance, the hand posture is changeable. Compared with the upper part of the human body, the small change of the hand posture is relatively easy to be ignored by the computer. The monocular camera recognition based on the traditional convolution neural network generally has the problems of low recognition accuracy and poor robustness. In addition, as a perceptual animal, people will show some emotions when communicating. These emotions are closely related to the meaning of sign language in the process of communication. This emotion is most directly reflected on the face, that is, facial expression. However, the facial expression containing effective information is often ignored in the general sign language recognition methods. 

Based on the above problems, we propose a new sign language recognition network that integrates hand skeleton features and facial expression features. The network inputs hand skeleton features and facial expression features into the network by means of data compensation. The improved skeleton feature extraction of the hand improves the precision of describing the spatial posture of the hand by using coordinate transformation and reduces the interference caused by different positions and angles of the hand in space. Self-Cure Network (SCN)~\cite{Kai_Wang} are used to extract features of facial expression that are often overlooked. A good use of the information can make sign language recognition more accurate and improve the robustness of sign language recognition methods.

\section{Related Work}
Dr. M. Madhiarasan and Prof. Partha Pratim Roy~\cite{Madhiarasan} reviewed the technical development of sign language recognition in the past decade and discussed the problems and limitations in the field of sign language recognition and the future development direction. For example, the model is highly dependent on the environment, difficult to adapt to complex changes in the environment, and poor in robustness. The singleness of the datasets and the lack of data volume lead to the fragility of the sign language recognition model. Therefore, standardized datasets with manual and non-manual characteristics are very important.

The skeleton can simply and clearly express the features related to the sign language action. We integrate the hand skeleton features into the sign language recognition, so that the model can resist the interference caused by different environments. De Smedt et al. ~\cite{De_Smedt} proved that the skeleton-based method is more accurate than the depth-based method on a challenging dataset through a series of comparative tests, but it lacks practical application in sign language. On this basis, Ran Cui et al.~\cite{Ran_Cui} proposed a multi-source motion recognition model combining the characteristics of time and space domains. The motion recognition based on the 3D skeleton information obtained from RGB-D videos has significantly improved in the recognition of a variety of general actions and interactive activities, but it is inappropriate to apply in general life scenes due to the high requirements of its equipment. For video analysis between consecutive frames, RNN has irreplaceable advantages. Kyunghyun Cho et al.~\cite{Cho_Kyunghyun} proposed a new neural network model of RNN coder-decoder composed of two recurrent neural networks, and proved the advantages of RNN in video analysis.

Sarfaraz Masood et al.~\cite{Masood_Sarfaraz} applied the combination of CNN and RNN to sign language recognition, which can only recognize sign language actions in a specific environment, that is, with poor robustness. In references~\cite{Jun_Liu}, ~\cite{Qinkun_Xiao}, skeleton information is applied to sign language recognition. In these works, the skeleton data is not standardized and the effective features of the data are not further refined. Based on this, we propose a method based on coordinate transformation, which can more accurately describe the hand posture corresponding to the sign language action.

In addition, there is relatively little research work on integrating facial expression features into sign language recognition. Facial expression can be used to assist sign language expression, so it has profound research significance in sign language recognition. Kai Wang et al.~\cite{Kai_Wang} proposed a Self-Cure Network to suppress the existing uncertainty. Based on the SCN, we realize the facial expression recognition of sign language actors, and integrate facial expression recognition into sign language recognition.

\section{Method}
Our work mainly focuses on isolated sign language vocabulary. Firstly, We extract 101 frames from each video sequence, convert each frame into a image and extract relevant data that can represent hand skeleton nodes and facial expression. Input the image of the hand obtained from the video sequence into Intercept-v3 model (provided by Tensorflow), and splice the output of Intercept-v3 model with the relevant data of the hand skeleton nodes and facial expression, then input it into the full connection layer. After that, input the output of the full connection layer into the LSTM model to learn the characteristics of the time series, and finally get the corresponding meaning of the sign language video sequence.

\subsection{Data Preprocessing}
We first draw the approximate area of the hand in a frame through the Mediapipe~\cite{Mediapipe} (An API provided by Google), turn the image outside the area into black, and then input the image of the hand into the follow-up network through threshold segmentation and mask according to the color characteristics, so as to reduce the color interference outside of hand contour.
\subsection{Skeleton Data Extraction Based on Coordinate Transformation}
\begin{figure}[h]
\centering\includegraphics[width = 4cm ]{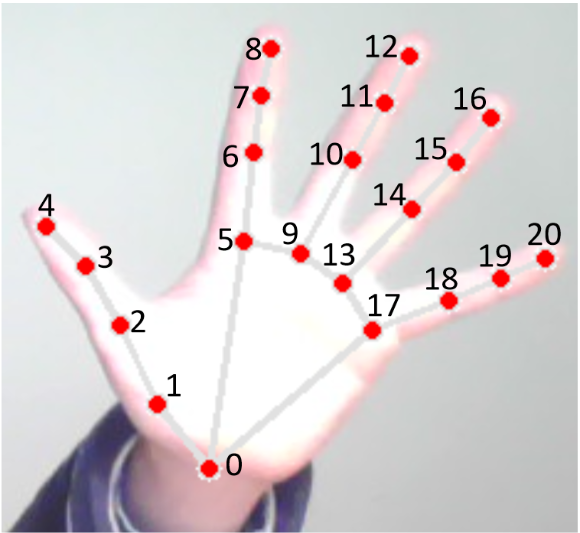}
\caption{Mediapipe hand joint points' corresponding serial numbers.} \label{Fig1}
\end{figure}

The human hand can be regarded as a rigid articulated system articulated by 21 joint points, so the information can be obtained from the coordinate of each joint point to describe the hand posture. We extract the 3D coordinates of the 21 joint points (see
Fig.~\ref{Fig1}) of the hand through Mediapipe from the RGB image after the frame extraction.

\begin{figure}[h]
\centering\includegraphics[width=0.49\textwidth]{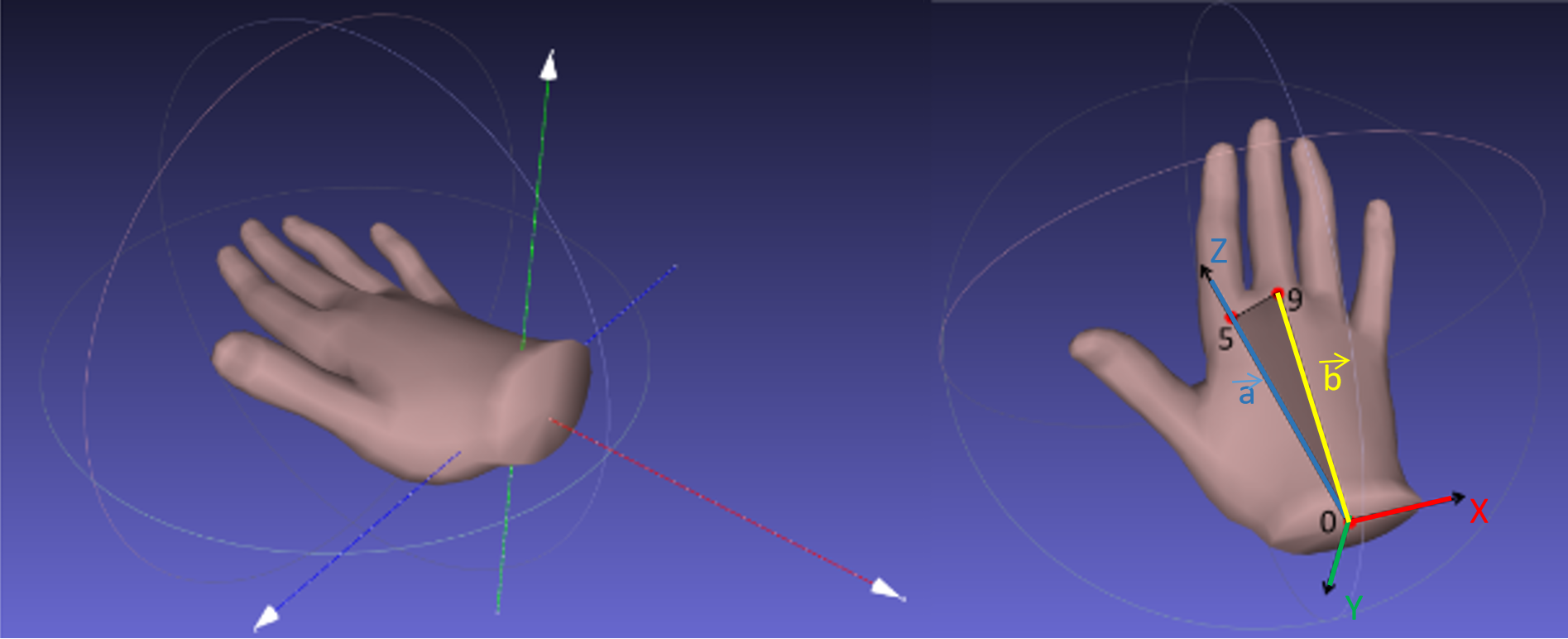}
\caption{ \centering Transformation from world coordinate system (left)
to \newline hand coordinate system (right).} 
\label{Fig2}
\end{figure}

The constructed neural network can distinguish the holistic position of hand in the picture, that is, the holistic position of hand in space, but it is difficult to accurately extract the gesture information of hand. Therefore, the recognition accuracy of hand posture is the key to determine the accuracy of sign language recognition. We calculate the relative coordinates of the obtained joint points' coordinates, then obtain the relative coordinates under the hand coordinate system through coordinate transformation (see
Fig.~\ref{Fig2}) to screen out its posture information, that is, retaining the relative position information of 21 joint points. This can reduce the impact of different hand orientations and different overall positions, and highlight the information of hand joint posture.

The processing method of coordinate transformation is as follows.

There are roughly two steps: translation and rotation. Because the hinge joints between the palm of the hand and the fingers except the thumb are basically inactive, it is difficult to move by themselves without the action of external forces. As shown in Fig.~\ref{Fig1}, nodes with serial numbers of 5, 9, 13 and 17 cannot move when the wrist joint point does not rotate. Therefore, the hand coordinate system (the right part of the Fig.~\ref{Fig2}) can be established based on this to accurately describe the hand posture.

The specific process of establishing the hand coordinate system are as follows. Make two vectors $\vec{a}$ and $\vec{b}$ from the skeleton node of the wrist to the hinge points of the index finger, the middle finger and the palm (nodes 5 and 9 in Fig.~\ref{Fig1}). Let these two vectors do the cross product of vectors and then we can get the $\vec{Y}$, which is perpendicular to the palm plane, that is, the shadow plane on the right side of Fig.~\ref{Fig2}. Let the direction of $\vec{a}$ be the positive direction of the Z coordinate axis in the hand coordinate system. Let $\vec{Y}$ and $\vec{a}$ do the cross product of vectors and then we can get the $\vec{X}$, which is the positive direction of the coordinate axis X under the hand coordinate system. According to the directions of $\vec{X}$, $\vec{Y}$, $\vec{a}$, a three-dimensional space rectangular coordinate system is established, which is the hand coordinate system.

The specific translation process is as follows. The coordinates in the world coordinate system of 21 joint points obtained via Mediapipe are $D_{i}(x_{i},y_{i},z_{i})$, where $D_{0}(x_{0},y_{0},z_{0})$ is the joint point at the wrist and i is the serial number representing the $i^{th}$ joint point, which is from 0 to 20 (see
Fig.~\ref{Fig1}). With $D_{0}$ as the coordinate origin, the relative wrist position coordinates of the 21 joint points of the hand are: $D_{i}(x_{i}-x_{0},y_{i}-y_{0},z_{i}-z_{0})$. At the same time, in order to minimize the interference of different hand scales on machine learning, the coordinates of hand joint points are normalized. When the scale is normalized, the hand posture is retained, that is, the relative position relationship of the spatial dimensions of each joint point remains unchanged. The vector length from the wrist to the hinge of the index finger and palm (node 5 in Fig.~\ref{Fig1}) is used as the standard for normalization. The normalized coordinate of each node after processing is as follows.
\begin{align}
    D_{i}(\frac{x_{i}-x_{0}}{\sqrt{(x_{5}-x_{0})^2+(y_{5}-y_{0})^2+(z_{5}-z_{0})^2}},\notag\\
    \frac{y_{i}-y_{0}}{\sqrt{(x_{5}-x_{0})^2+(y_{5}-y_{0})^2+(z_{5}-z_{0})^2}}, \notag\\
    \frac{z_{i}-z_{0}}{\sqrt{(x_{5}-x_{0})^2+(y_{5}-y_{0})^2+(z_{5}-z_{0})^2}})
\end{align}

The above results are only the normalized coordinates obtained after translation. Rotation refers to the rotation of the hand coordinate system relative to the world coordinate system. According to this rule, the normalized coordinates can be converted into the coordinates under the hand coordinate system.

The essence of rotation coordinate transformation is projection. Let $\overrightarrow{OM}=(p,q,r)$ be a unitized vector in the hand coordinate system (point O is the original point of the hand coordinate system), then project this vector to the world coordinate axis and accumulate it in the direction to obtain the rotation matrix R. Multiply the rotation matrix left by $\overrightarrow{OM}$ to get the coordinates of the M point in the world coordinate system. Then the coordinate corresponding relation before and after coordinate system transformation is as follows.
\begin{align}
    \begin{bmatrix}
 x\\
 y\\
 z
\end{bmatrix} =R\times \begin{bmatrix}
 p\\
 q\\
 r
\end{bmatrix}=\begin{bmatrix}
  p_{x}&q_{x}&r_{x}\\
  p_{y}&q_{y}&r_{y}\\
  p_{z}&q_{z}&r_{z}
\end{bmatrix}\times \begin{bmatrix}
 p\\
 q\\
 r
\end{bmatrix}
\end{align}

Similarly, the transposition of matrix R is the rotation matrix from the world coordinate system to the hand coordinate system. With the help of the rotation matrix, the coordinates of 21 hand joint points in the hand coordinate system can be obtained. $D_{i}^{'}$ is the final coordinate of the $i^{th}$ joint point in the hand coordinate system after coordinate transformation.
\begin{align}
    \overrightarrow{OD_{i}^{'}}=R^{T}\times  \overrightarrow{OD_{i}}
\end{align}

Through the above coordinate transformation, we can get the coordinates of 21 standardized joint points under the hand coordinate system. Each frame corresponds to a set of standardized joint point coordinates under the hand coordinate system, which will be input into the sign language recognition network to assist the network to recognize the sign language posture in the image.
\subsection{Facial Expression Data Extraction}
Facial expression data extraction is based on SCN proposed by Wang et al.~\cite{Kai_Wang} This network uses self-attention importance weighting, ranking regularization and relabeling to suppress the uncertainty in facial expression recognition. We collect 26k images from the Internet containing a wide variety of facial expression, crop the facial area through OpenCV, and extract a square area that can cover all feature points of the face. The extracted images are converted into 48×48 gray-scale images for training the network, and finally a model is obtained that can recognize 7 kinds of expressions: angry, disgusted, fear, happy, neutral, sad and surprised. The model is used to predict the facial expression of each frame of a sign language video sequence, and the confidence of each frame in each expression category is extracted, which will be helpful in sign language recognition. These data will be used for the subsequent input of sign language recognition network to assist the network to identify the sign language gestures in the image.
\subsection{Data Compensation Network}
We input the skeleton data and facial expression data obtained in the previous two sub-steps into the network to enhance the accuracy and reliability of sign language recognition. Since a large number of sign language information has both temporal and spatial characteristics, we combine convolutional neural network (CNN) and recursive neural network (LSTM) in terms of network. Convolutional neural network is used to learn the spatial features of images, and recursive neural network is used to learn the changes of hand posture and facial expression in time series.

CNN can effectively screen out the effective information in the image and classify the image according to the spatial features. The convolution layer of CNN can effectively extract the effective features of the image data and screen the general position and shape information of the hand in the image. The pool layer of CNN can further remove information that is not useful for learning sign language. Finally, the basic image classification task can be completed through the weight calculation of the full connection layer. RNN can capture continuous time features, learn a video sequence's changes of a sign language word, discover changes of hand postures in time, and finally complete video sequence classification.
\begin{figure}[h]
\centering\includegraphics[width=0.47\textwidth]{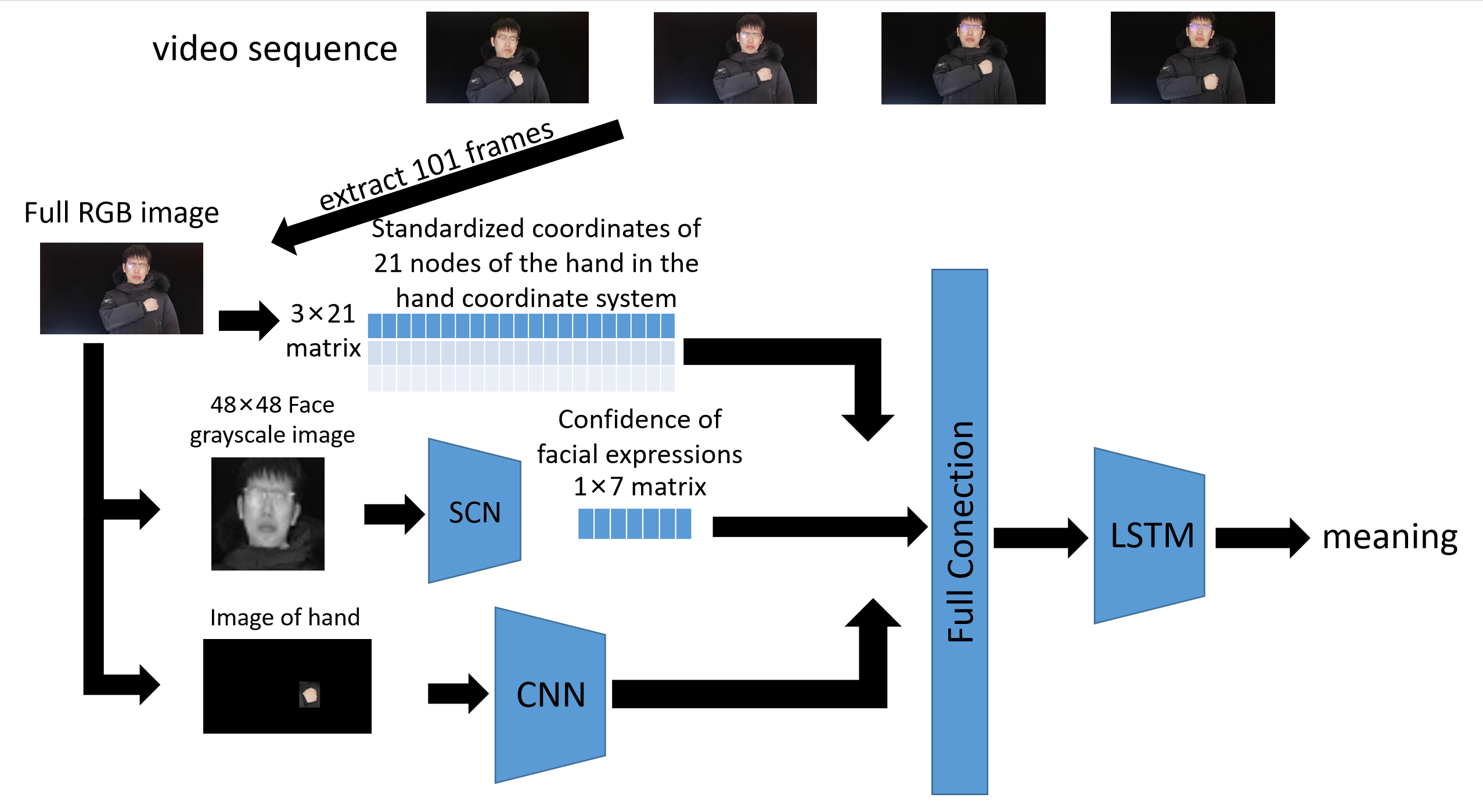}
\caption{ \centering New sign language recognition data compensation network} 
\label{Fig3}
\end{figure}

We add the data related to the hand skeleton and facial expression, so that the network can obtain more effective information conducive to sign language recognition. As shown in Fig.~\ref{Fig3}, we extract the three-dimensional coordinate information of the hand skeleton in a single image and carry out coordinate transformation to obtain the relative coordinate information of the $21\times 3$ array. Based on SCN which can suppress uncertainty of facial expression recognition proposed by Wang et al., confidence information of each image face in each expression category is obtained, namely, confidence array with length of 7 (that is, the number of expression classes). A valid information array of $21\times 3+7$ can be obtained by combining the coordinate information of hand and the confidence information of facial expression in a single image. CNN will perform category recognition on a single image, but its category recognition is often difficult to effectively screen the information of hand gesture changes, and it is easy to ignore the facial expression information, which is conducive to sign language recognition. Therefore, we can combine the extracted $21\times 3+7$ effective information array with CNN network to improve the recognition ability of gesture changes of the hand and facial expression, so as to realize more effective sign language recognition. Given that this array of valid information is already quite refined, there is little need to sift out redundant information further. Therefore, we directly splice the effective information array with the CNN output array with the length of 1024, and then input into the full connection layer for classification. Finally, more accurate confidence information of each picture in each sign language category can be obtained, namely an array of length n (n is the number of sign language classes). After the class confidence array with length n of each frame is obtained, the confidence array belonging to the same sign language video is packaged as the input of LSTM for training and testing of LSTM.

\section{Experiments}
\subsection{Sign language dataset}
 As is known to all, the production of video-based sign language datasets is relatively difficult, resulting in a relatively slow development of sign language recognition. We collected the Chinese video sequence sign language recognition dataset called SEUCSLRD, and the meaning of each sign language video corresponds to a word. The dataset was recorded by four signers, and 20 commonly used Chinese sign languages were selected for recording (specific categories are shown in Table~\ref{tab1}). The same signer records the same movement for multiple times, effectively avoiding the problem of non-standard sign language movements in a single recording. Each video is about 2 seconds long.
 
 The recording of this dataset was strictly regulated. The video recording device was required to be about 1.5 meters in front of the signer, and horizontal screen shooting was adopted. The signers kept the background and clothing black and let the clothing to cover the neck and arms, showing the hands and face of the signer, which is the most effective part of the sign language movement. In addition, for the convenience of researchers, the dataset was preprocessed to cut out invalid frames such as background frames at the beginning and end of the videos. Unlike other datasets, this dataset emphasizes the need for signers to record facial expressions of the corresponding sign language words.
\begin{table}[h]
\caption{SEUCSLRD dataset description (* represents that the sign language action has obvious facial expression)}\label{tab1}
 \setlength{\tabcolsep}{0.7mm}{
\begin{tabular}{|ccc|ccc|}
\hline
\multicolumn{1}{|c|}{\textbf{Category No.}} & \multicolumn{1}{c|}{\textbf{Meaning}} & \textbf{Expression} & \multicolumn{1}{c|}{\textbf{Category No.}} & \multicolumn{1}{c|}{\textbf{Meaning}} & \textbf{Expression} \\ \hline
1                                           & angry            & *                   & 11                                         & laugh                                 &                     \\ \hline
2                                           & attention                             &                     & 12                                         & pull                                  &                     \\ \hline
3                                           & bad                                   &                     & 13                                         & rabbit                                &                     \\ \hline
4                                           & disgust                               & *                   & 14                                         & sad                                   & *                   \\ \hline
5                                           & electricity                           &                     & 15                                         & simple                                &                     \\ \hline
6                                           & fake                                  &                     & 16                                         & smile                                 & *                   \\ \hline
7                                           & fear                                  & *                   & 17                                         & strive                                &                     \\ \hline
8                                           & fifty                                 &                     & 18                                         & unhappy                               & *                   \\ \hline
9                                           & good                                  & *                   & 19                                         & water                                 &                     \\ \hline
10                                          & high                                  &                     & 20                                         & you                                   &                     \\ \hline
\end{tabular}
}
\end{table}

We also conducted tests on the famous Argentine Sign Language Dataset (LSA64~\cite{Ronchetti2016} ), which consists of 3,200 videos of 10 non-expert subjects performing five different types of sign language for 64 times for each type, using the most commonly used symbols in Argentine sign language.

\subsection{Experimental analysis and discussion}
 We compared the data compensation network including hand coordinates and facial expression proposed by us with the ordinary CNN-LSTM network that directly input image information. In addition, we observed the effect of adding hand skeleton features or facial expression features alone through a comparative experiment. These methods used the SEUCSLRD and LSA64 datasets for training and testing. This experiment is mainly to compare and clarify the positive role of hand skeleton feature extraction based on coordinate transformation and facial expression features in sign language recognition. First of all, the training dataset and the testing dataset were divided according to the ratio of 1 to 4.

 In these methods, the frame extracted from the training dataset is handed over to the CNN model for spatial feature training, and then the obtained model is used to predict and store the frame extracted from the training dataset and the testing dataset, and then the CNN prediction data corresponding to the training dataset is handed over to the LSTM model for temporal feature training. Finally, after training the LSTM model, the CNN prediction data corresponding to the testing dataset is input into the LSTM model for testing. The size of each batch of the two methods was set to 100, and the number of training epochs of the LSTM model was set to 100.

 Finally, after testing, in the SEUCSLRD data set, the recognition accuracy of ordinary CNN-LSTM network was 72.5$\%$, the recognition accuracy of our proposed data compensation network was 92$\%$, the network accuracy of facial features added alone was 80$\%$, and the network accuracy of hand skeletal features added alone was 82.5$\%$. In the LSA64 dataset, the recognition accuracy of the ordinary CNN-LSTM network was 87.5$\%$, and the accuracy of the network adding facial features alone was 91.25$\%$. The reason for the lack of the comparative experiment related to the extraction of hand skeleton features in the LSA64 dataset is that it is difficult to obtain hand skeleton features by ordinary methods, such as Mediapipe, since the shooting of LSA64 dataset was done with gloves.
 \begin{figure}[h]
\centering\includegraphics[width=0.37\textwidth]{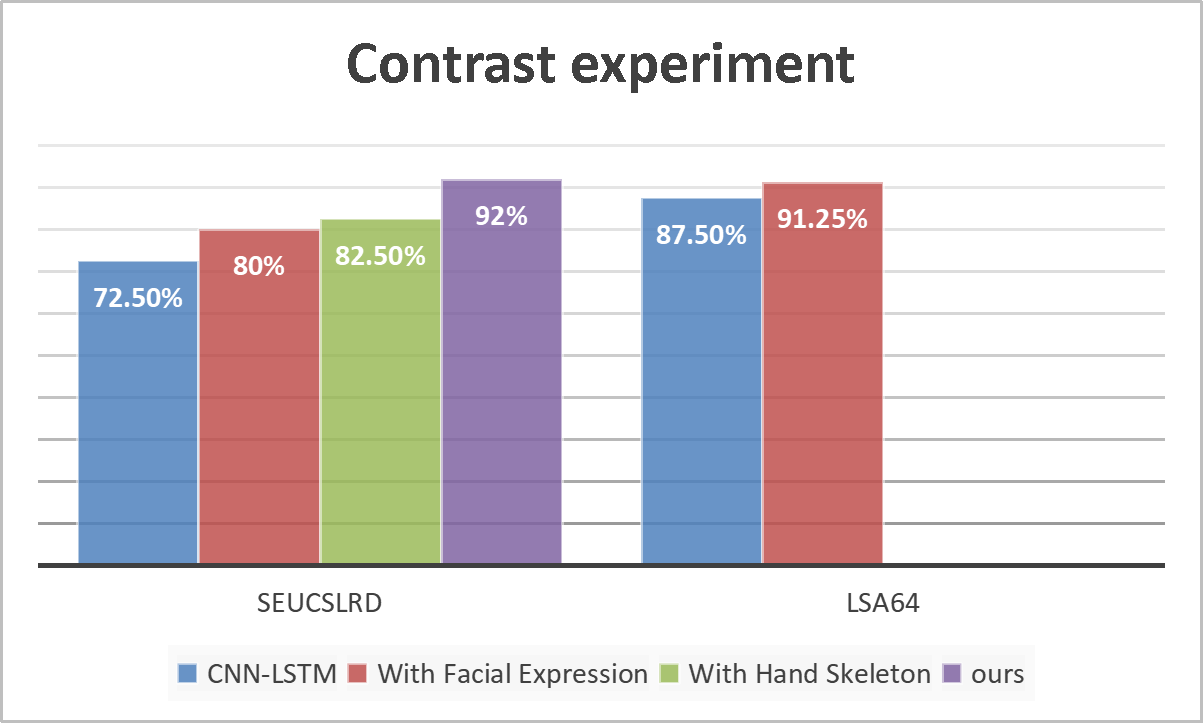}
\caption{ \centering Comparison of accuracy of different methods on SEUCSLRD and LSA64 datasets} 
\label{Fig4}
\end{figure}

 It can be found that the data compensation network has better sign language recognition ability, which improves its performance. This method of integrating hand skeleton features to obtain more effective hand posture information and rich facial expression features to assist sign language recognition is very effective, which is of enlightening significance for future research.

\section{Conclusion}
In this paper, we propose a data compensation network that integrates skeleton features of the hand and facial expression features, which can improve the accuracy and robustness of sign language recognition. The extraction of hand skeleton features is based on coordinate transformation, which can obtain hand posture more accurately and avoid interference caused by different position and angle of the hand in space. The SCN can extract facial expression information neglected in a large number of sign language recognition methods, which assists in sign language recognition. A large number of comparative experiments done on a public dataset LSA64~\cite{Ronchetti2016} and SEUCSLRD collected by us show that our data compensation network integrating hand skeleton features and facial expression features has achieved very good results, which is encouraging to effectively process the data under different interference.

\section*{Acknowledgment}
Thanks to Prof. Yangang Wang and Dr. Zimeng Zhao from the School of Automation, Southeast University for their guidance.

\vspace{12pt}

\end{document}